\pdfoutput=1
\documentclass{article}

\PassOptionsToPackage{numbers,compress}{natbib}
\usepackage[preprint]{neurips_2026}

\usepackage[utf8]{inputenc}
\usepackage[T1]{fontenc}
\usepackage{hyperref}
\hypersetup{hypertexnames=false}
\usepackage{url}
\usepackage{booktabs}
\usepackage{amsfonts}
\usepackage{nicefrac}
\usepackage{xcolor}
\usepackage{microtype}
\usepackage{amssymb}
\usepackage{amsmath}
\usepackage{algorithm}
\usepackage{algorithmic}
\usepackage{multirow}
\usepackage{multicol}
\usepackage{colortbl}
\usepackage{graphicx}
\usepackage{enumitem}
\usepackage{makecell}
\usepackage{tabularx}
\usepackage[breakable]{tcolorbox}

\title{MoCA-Video: Motion-Aware Concept Alignment for Video Semantic Mixing}

\author{%
  Tong Zhang$^{1}$ \quad
  Victor Escorcia$^{2}$ \quad
  Juan Carlos Leon Alcazar$^{1}$ \quad
  Bernard Ghanem$^{1}$\\[0.6em]
  $^{1}$King Abdullah University of Science and Technology\\
  $^{2}$MacondoLabs
}

\begin{document}
\maketitle

\begin{abstract}

Unlike traditional video editing or inpainting, video semantic mixing fuses a reference concept with a moving target entity to produce a hybrid while preserving the source video's motion and layout.
We propose MoCA-Video, a training-free framework that steers a frozen video-diffusion denoising trajectory through concept-localized reference injection.
At selected low-noise steps, MoCA-Video uses concept attention to localize the target object and injects the reference latent into the localized region, where object structure has formed but appearance remains editable.
A momentum-based correction carries the injected prediction across frames to encourage coherent concept integration through the sequence.
We further introduce CASS, a CLIP-based metric that measures the output's directional alignment shift toward the reference and away from the source prompt.
Using the denoiser's internal attention avoids an external localization model; in our A100 FP16 setup, MoCA-Video takes 3.2 seconds per output frame, excluding preprocessing.
Across the evaluated baselines, MoCA-Video achieves the highest CASS, rel-CASS, and ImageReward, while LPIPS-T and FVD expose separate temporal-coherence and video-quality trade-offs.

\end{abstract}

\section{Introduction}
\label{sec:intro}

Video diffusion models~\citep{ho2022video, hong2022cogvideo, kong2024hunyuanvideo, opensora} must synthesize temporally coherent frames in addition to denoising images.
Recent work~\citep{singer2022makeavideotexttovideogenerationtextvideo, chen2024videocrafter2, wan2025, kong2024hunyuanvideo, chen2023pixartalphafasttrainingdiffusion} improves visual quality and temporal consistency.
These advances support image-to-video animation~\citep{xing2023dynamicrafter, guo2023animatediff}, subject-driven editing~\citep{ku2024anyv2vtuningfreeframeworkvideotovideo, huang2024divetamingdinosubjectdriven,wei2024dreamvideo2zeroshotsubjectdrivenvideo, kim2025subjectdrivenvideogenerationdisentangled}, and affordance insertion~\citep{kulal2023affordance, he2024affordance}, among other applications.
However, the training data constrain which entity combinations these models can readily synthesize and manipulate.
This limitation is evident in the \textit{video semantic mixing} task, whose goal is to produce temporally coherent hybrid objects that preserve key structural properties of the original subject while incorporating complementary semantics from a reference source.

\begin{figure*}[t]
  \centering
  \includegraphics[width=0.95\linewidth]{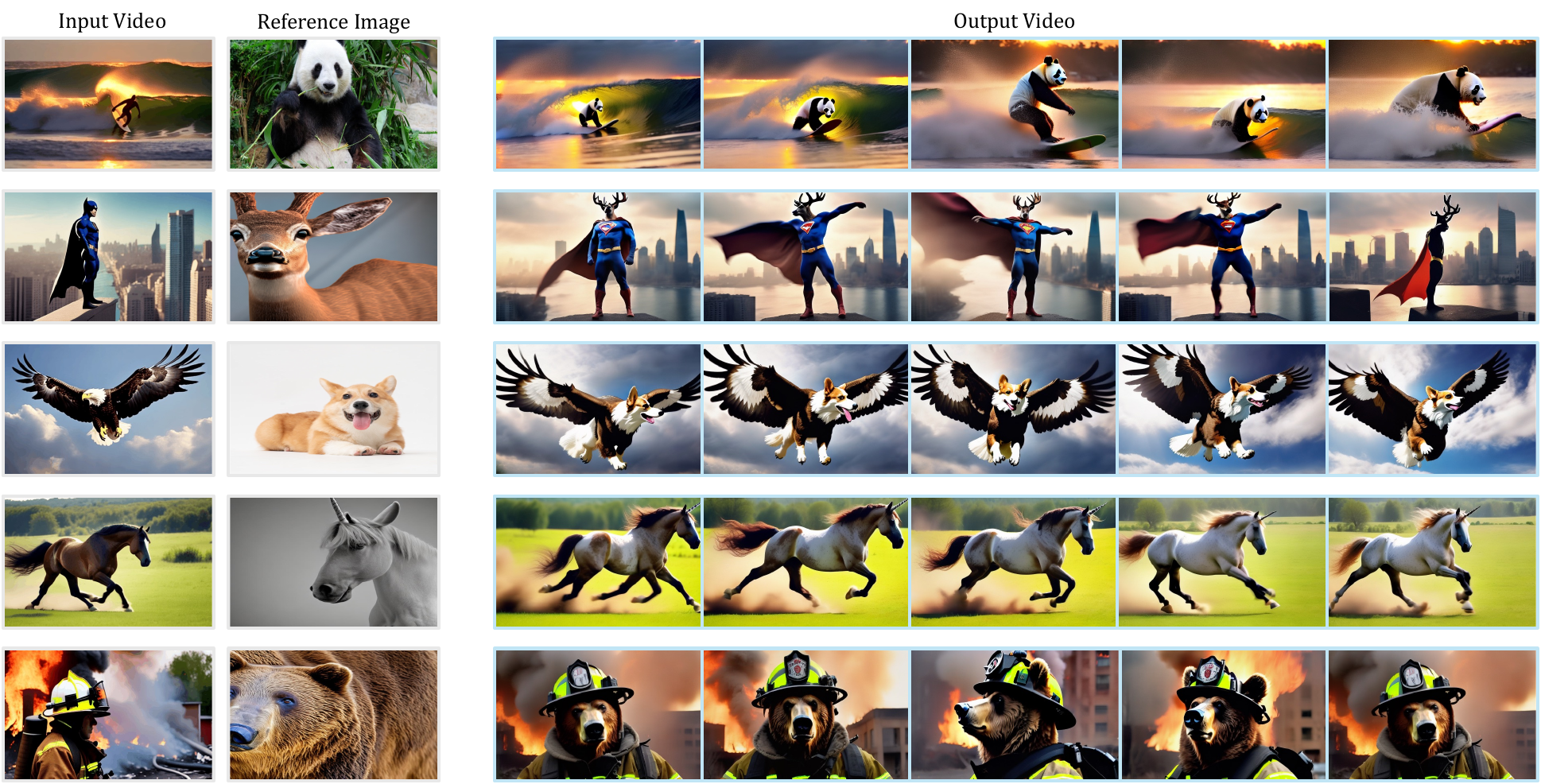}
  \vspace{-5pt}
  \caption{\textbf{Video semantic mixing with \modelnameabbr.} Given a video and reference image, \modelnameabbr\ blends their concepts across diverse object categories. Prompts are provided in the Supplementary Material.}
  \label{fig:semantic_mixing_examples}
  \vspace{-15pt}
\end{figure*}

While semantic mixing has been explored in images~\citep{liew2022magicmix,zhou2025freeblend}, extending it to video requires localized control and temporal consistency.
Existing frame-wise, style-transfer, and conditioning-based methods often entangle local and global features.
We introduce MoCA-Video (Motion-Aware Concept Alignment in Video), a training-free framework that injects reference semantics into targeted entities by manipulating latent noise trajectories to produce temporally coherent hybrid concepts (Fig.~\ref{fig:semantic_mixing_examples}).
MoCA-Video operates entirely at inference time on a frozen video diffusion model, providing an empirical study of structured control during iterative denoising.

To evaluate video semantic mixing, we introduce the Concept Text--Video Blending (CTVB) dataset and benchmark, which extends the Concept Text--Image Blending (CTIB) dataset~\citep{zhou2025freeblend} with DAVIS entities~\citep{Perazzi2016}, 21 base videos, and 106 intra- and inter-category video--image pairs.
CTVB measures directional concept alignment with CASS/rel-CASS, temporal coherence with LPIPS-T~\citep{zhang2018unreasonableeffectivenessdeepfeatures}, and video quality with FVD and ImageReward.
MoCA-Video achieves the highest CASS, rel-CASS, and ImageReward in our comparison, although several baselines obtain lower LPIPS-T or FVD.
Our contributions are:

\begin{itemize}
    \item \noindent\textit{The MoCA-Video framework.} We introduce a training-free framework for video semantic mixing that fuses a reference concept with a moving target object in a frozen video diffusion model.
    MoCA-Video couples concept-attention localization, which restricts direct injection to the target region, with a momentum-based correction that carries the injected prediction across frames.
    \item \noindent\textit{The CTVB benchmark.} We extend CTIB with 106 intra- and inter-category video--image pairs and evaluate them with CASS/rel-CASS alongside temporal, video-quality, qualitative, and human measures.
\end{itemize}
We compare MoCA-Video with training-free and pretrained video editing methods and ablate its components and principal hyperparameters.
The concept-attention formulation also avoids external segmentation or tracking and takes 3.2 seconds per output frame in our reported A100 FP16 setup, excluding preprocessing.

\section{Related Work}
\label{sec:formatting}

\subsection{Semantic Mixing and Concept Blending}
Semantic mixing has been studied in the image domain through MagicMix~\citep{liew2022magicmix} and later FreeBlend~\citep{zhou2025freeblend}, both of which blend disparate concepts into novel objects. These methods exploit the denoising dynamics of diffusion models to factorize layout and content or apply staged latent interpolation for more stable blending. However, both approaches are limited to static images and do not address temporal consistency. MagicEdit~\citep{liew2023magicedit} extends semantic editing to video by injecting prompt-driven features at specific diffusion stages. While it preserves motion to some degree, it lacks spatial control and explicit fusion of image-based semantics.

\subsection{Identity-Preserved Video Generation}
A closely related line of work is identity-preserved text-to-video generation (IPT2V), which aims to retain a reference subject's appearance while synthesizing novel motion. ID-Animator~\citep{he2024idanimatorzeroshotidentitypreservinghuman} enables zero-shot face-driven video generation but tends to overfit to the input image. ConsisID~\citep{yuan2024identitypreservingtexttovideogenerationfrequency} and EchoVideo~\citep{wei2025echovideoidentitypreservinghumanvideo} further enhance facial detail through frequency-aware control and multimodal fusion, respectively, yet remain restricted to human identities. More recently, MagicMirror~\citep{zhang2025magicmirror} introduces a dual-branch facial feature extractor with conditional adaptive normalization for DiT-based architectures, achieving stronger identity consistency but still operating exclusively on human faces. These IPT2V methods primarily focus on preserving a reference identity during generation, whereas video semantic mixing seeks to fuse reference characteristics with an existing video entity to produce a hybrid combining attributes from both sources.

\subsection{Video Inpainting}
Video inpainting methods extend region-specific filling to the temporal domain, typically relying on optical flow, feature correspondences, or learned priors to maintain frame-to-frame consistency. DIVE~\citep{huang2024divetamingdinosubjectdriven} leverages DINO features with LoRA-based identity registration for subject-driven edits, while ObjectMate~\citep{winter2024objectmaterecurrencepriorobject} builds an Object Recurrence Prior trained on large-scale supervised compositions. TokenFlow~\citep{tokenflow2023} propagates self-attention tokens via nearest-neighbor matching to enforce smooth transitions, and RAVE~\citep{kara2023raverandomizednoiseshuffling} shuffles latent and condition grids during denoising to preserve consistency upon unshuffling. More recently, VideoPainter~\citep{bian2025videopainter} extends diffusion-based inpainting to arbitrary-length videos with improved temporal coherence. While inpainting methods can achieve a degree of concept mixing by replacing masked regions with prompt-guided content, they tend to fully overwrite the original appearance rather than blend it with new semantics, resulting in identity discontinuities and degraded video quality.

\section{Methodology}

\subsection{Problem Formulation}
Given a source video $V^{src}=\{I^{src}_1,\ldots,I^{src}_F\}$, a source prompt $p^{src}$, a target concept $c^{src}$, and a reference image $I^{ref}$ representing a conditioning concept $c^{ref}$, our goal is to generate a fused video $V^{mix}$ that preserves the source layout and motion while transforming the target entity toward a hybrid concept.
Formally, the output should satisfy:
\begin{align}
    \mathrm{layout}(V^{mix}) &\approx \mathrm{layout}(V^{src}), \\
    \mathrm{motion}(V^{mix}) &\approx \mathrm{motion}(V^{src}), \\
    \mathrm{semantics}(V^{mix}_{target}) &\approx \mathrm{hybrid}(c^{src}, c^{ref}).
\end{align}
The target entity should retain the source's spatial structure and temporal dynamics while taking on visual semantics and affordances from the reference concept.
This objective seeks a localized fusion within a moving object rather than a text-specified edit or masked-region replacement.
\subsection{Overview}

\modelnameabbr\ edits an input video using a reference image $I^{ref}$ entirely within the latent trajectory of a frozen text-to-video diffusion model.
Given the noisy video latent $x_t \in \mathbb{R}^{F \times C \times H \times W}$ and reference latent $z^{ref}=E(I^{ref})$, the method edits during low-noise steps ($t \leq t'$), when object structure is largely established but appearance remains malleable.

\begin{figure*}[t]
  \centering
  \includegraphics[width=0.9\textwidth]{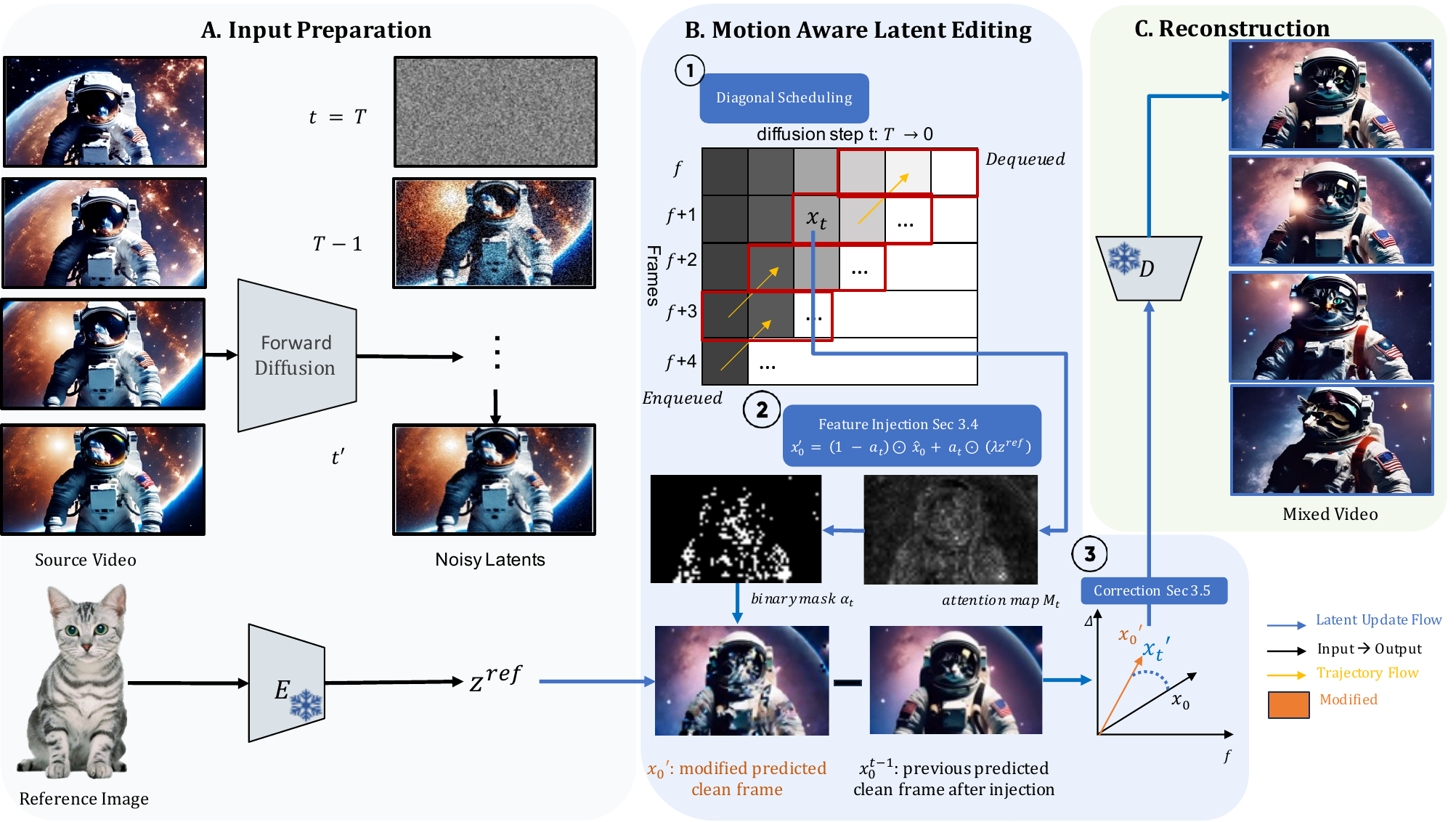}
  \caption{\textbf{Overview of \modelnameabbr.} Given a video and reference image, we inject $z^{\mathrm{ref}} = E(I^{\mathrm{ref}})$ into concept-attention regions of $\hat{x}_0$ during low-noise steps ($t \leq t'$) (Sec. \ref{sec:concept_localization}), use diagonal scheduling and momentum correction for bounded cross-frame propagation (Sec. \ref{sec:momentum}), and decode the edited latents with frozen decoder $D$.}
  \label{fig:method}
\end{figure*}

Built on a diagonal denoising scheduler~\citep{kim2024fifo}, \modelnameabbr\ localizes the target concept $c^{src}$ from cross-attention, injects $z^{ref}$ into the resulting region of the predicted clean sample $\hat{x}_0$, and carries the edit across frames using momentum-based correction.
These operations produce the localized blend shown in Fig.~\ref{fig:method} without external segmentation, tracking, or frame-wise editing.

\subsection{Video Diffusion Preliminary}
\label{sec:preliminary}
Video diffusion models learn to reverse a gradual noising process to generate temporally coherent frame sequences.
Given a video $\mathbf{x}_0 \in \mathbb{R}^{T \times C \times H \times W}$, the forward process progressively adds Gaussian noise over $T$ timesteps according to a variance schedule $\{\alpha_t\}_{t=1}^{T}$, where $\alpha_t$ denotes the cumulative product up to step $t$:
\begin{equation*}
q(x_t \mid x_0) = \mathcal{N}\bigl(x_t;\, \sqrt{\alpha_t}\,x_0,\, (1 - \alpha_t)\,\mathbf{I}\bigr).
\end{equation*}
A denoising network $\epsilon_\theta$ predicts the added noise, and the reverse process recovers $x_0$ from $x_T \sim \mathcal{N}(\mathbf{0}, \mathbf{I})$.
Following DDIM~\citep{song2022denoisingdiffusionimplicitmodels}, each reverse step predicts the clean sample
\begin{equation*}
\hat{x}_0^{(t)} = \frac{x_t - \sqrt{1-\alpha_t}\,\epsilon_\theta(x_t,t)}{\sqrt{\alpha_t}},
\end{equation*}
and updates the latent to $x_{t-1}$ with an added stochastic term:
\begin{equation}
\label{eq:ddim_reverse}
\begin{aligned}
x_{t-1} &= \sqrt{\alpha_{t-1}}\,\hat{x}_0^{(t)} + d_t + \sigma_t\,\epsilon_t, \\
d_t &= \sqrt{1-\alpha_{t-1}-\sigma_t^2}\,\epsilon_\theta(x_t,t),
\end{aligned}
\end{equation}
Here, $d_t$ is the directional term and $\sigma_t$ sets the per-step stochasticity.
To edit an existing video, we encode it and apply the forward process above.
\modelnameabbr\ then intervenes during later, low-noise steps ($t \le t'$), where structural layout is largely established and appearance remains malleable.
Concept-attention injection modifies the predicted clean sample $\hat{x}_0^{(t)}$ within the localized fusion region, and momentum-based correction carries the injected prediction across frames.
Our experiments use the UNet-based VideoCrafter2 backbone.
Our approach acts on the latent space of a reverse diffusion process, and is independent of the particular $\epsilon_\theta$ parameterization.

\subsection{Concept-Localized Injection}
\label{sec:concept_localization}

\modelnameabbr\ localizes the target object at each denoising step using the model's cross-attention, enabling region-specific latent injection without external segmentation or tracking.

\noindent\textbf{Concept attention.}
Within the spatial cross-attention layers of the denoiser, queries $Q$ are projected from the spatial latent features and keys $K$ from the text-token embeddings of the source prompt.
The attention map is
\begin{equation}
\label{eq:attention}
    A = \mathrm{softmax}\!\left(\frac{Q K^\top}{\sqrt{d}}\right).
\end{equation}

We select the columns of $A$ corresponding to the token indices of the target concept $c^{src}$ (e.g., \emph{astronaut}) and aggregate them into a single spatial map.
We use only the conditioned branch of classifier-free guidance and restrict aggregation to spatial layers above a minimum resolution because the coarsest layers blur the localization.
For each retained layer, we average over attention heads and target-token indices to obtain a per-layer map.
We then bilinearly interpolate each map to the latent resolution and average the maps with weights proportional to layer resolution, so that finer layers dominate.
Min--max normalization maps the aggregate to $M_t \in [0,1]^{H_z \times W_z}$.
We threshold it at $\tau$ to obtain the binary localization mask
\begin{equation}
\label{eq:mask}
    a_t = \mathbb{1}[\,M_t > \tau\,],
\end{equation}
which localizes the target object at timestep $t$.

Because $M_t$ reuses attention from the denoiser's forward pass, localization does not require an external segmentation or cross-frame tracking model.
The map adapts as the object's appearance evolves.
Under the diagonal scheduler, frames within a window are processed at staggered noise levels in a single pass, so a freshly captured mask is shared across the window while varying across timesteps.

\noindent\textbf{Injection.} The binary mask $a_t$ acts as a spatial gate that restricts direct injection to the target region while leaving the surrounding latent unchanged at that operation.
Injection operates on the predicted clean sample $\hat{x}_0$ during low-noise steps $t \le t'$.
Within the gated region, we substitute reference content and leave the complement unchanged:
\begin{equation}
\label{eq:injection}
    x_0' = (1 - a_t)\odot \hat{x}_0 + a_t \odot \bigl(\lambda\, z^{ref}\bigr),
\end{equation}
Here, $z^{ref} = E(I^{ref})$ is the encoded reference latent, $\lambda$ is the conditioning strength, $\odot$ is element-wise multiplication broadcast over channels, and $x_0'$ is the modified clean prediction.
The mask confines direct substitution to the target region, although subsequent denoising can still change its complement.
Because injection occurs inside the denoising loop, the denoiser can update boundary transitions over subsequent steps.

The conditioning strength $\lambda$ controls the reference magnitude within the localized region: larger $\lambda$ amplifies the injected features.
We use a single fixed $\lambda$ for all pairs, selected by the ablation in Sec. \ref{sec:ablation}.
The injected prediction $x_0'$ does not alter the current step's output latent, which has already been formed.
Instead, the method carries $x_0'$ into the next frame's correction (Sec. \ref{sec:momentum}), allowing the blended concept to propagate under the diagonal schedule.

\subsection{Momentum-Based Concept Propagation}
\label{sec:momentum}

A momentum buffer carries localized reference injections across frames along the diagonal schedule.
Algorithm~\ref{alg:momentum_unified} in the Supplementary Material specifies the complete procedure.

\noindent\textbf{Correction.}
Let $\hat{x}_{0,f}^{(t)}$ be the raw clean prediction for frame
$f$ at diffusion timestep $t$, and let
$\tilde{x}_{0,f-1}^{(t)}$ be the retained prediction from the
preceding frame after injection.
We form a frame-difference signal and accumulate it in a velocity
buffer $v_{t,f}$ using an exponential moving average:
\begin{equation}
\label{eq:momentum}
\begin{aligned}
g_{t,f}
&= \bigl(\hat{x}_{0,f}^{(t)}
- \tilde{x}_{0,f-1}^{(t)}\bigr)
+ \eta\,d_{t,f}, \\
v_{t,f}
&= \mathrm{clip}\!\bigl(
\beta v_{t,f-1}
+ (1-\beta)g_{t,f},\,-c,\,c
\bigr).
\end{aligned}
\end{equation}
Here,
$d_{t,f}
= \sqrt{1-\alpha_{t-1}-\sigma_t^2}\,
\epsilon_\theta(x_{t,f},t)$
is the DDIM directional term for frame $f$,
$\eta=0.05$ weights its contribution,
$\beta=0.9$ controls the momentum,
and the buffer is clipped to $[-c,c]$ with $c=2.0$.
Because $\tilde{x}_{0,f-1}^{(t)}$ carries the reference injected
into the preceding frame, $g_{t,f}$ measures how the current raw
prediction differs from the preceding injected prediction.
The EMA retains components shared across adjacent frames while
attenuating frame-specific variation.

We correct the current prediction along the accumulated direction:
\begin{equation}
\label{eq:corrected}
\hat{x}_{0,f}^{\mathrm{(corr)},(t)}
= \hat{x}_{0,f}^{(t)}+\kappa_t v_{t,f},
\qquad
\kappa_t=\kappa_0(1-t/T).
\end{equation}
We use $\kappa_0=0.1$ and $T=1000$ and form the output latent by
applying the standard DDIM update to
$\hat{x}_{0,f}^{\mathrm{(corr)},(t)}$.
The correction strength grows as denoising proceeds toward $t=0$:
it is gentle at high noise and strongest during the late,
low-noise steps where appearance consolidates.

\noindent\textbf{Propagation along the diagonal.}
Injection modifies the clean prediction after the current output
latent has been formed.
The method retains the injected prediction as
$\tilde{x}_{0,f}^{(t)}$, which enters the next frame's correction
through $g_{t,f+1}$.
Under the diagonal schedule, consecutive frames occupy adjacent
positions along the noise diagonal.
This delayed coupling carries the reference signal from frame to
frame, while clipping bounds each individual correction.

\section{Experiments}
\label{sec:experiment}

We introduce the Concept Text--Video Blending (CTVB) dataset and benchmark, extending the image-based CTIB taxonomy to video--image pairs and a multi-axis evaluation of directional concept alignment, temporal coherence, video quality, and human judgments.
We use CTVB to compare training-free and pretrained baselines and to ablate \modelnameabbr's components and hyperparameters.

\subsection{Entity Blending Dataset}
\label{sec:dataset}

CTVB extends CTIB's four-super-category text--image taxonomy~\citep{zhou2025freeblend} to video semantic mixing by adding DAVIS-16 classes~\citep{Perazzi2016} organized into 16 subcategories.
It contains 21 base videos and 106 video--image pairs covering intra-category and semantically distant inter-category combinations (Fig.~\ref{fig:dataset_construction}).
Additional examples appear in the Supplementary Material.

\begin{figure*}[h]
  \vspace{-2pt}
  \centering
  \includegraphics[width=0.95\linewidth]{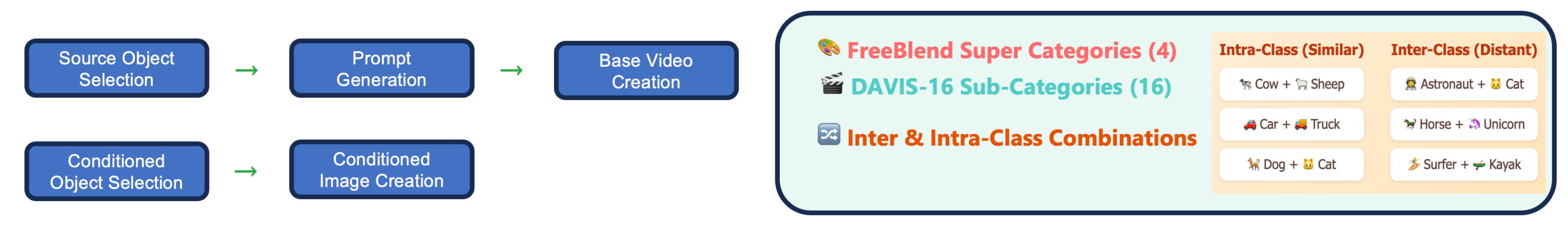}
   \vspace{-10pt}
  \caption{\textbf{Entity-blending dataset pipeline.} CTIB and DAVIS-16 entities are paired to create intra- and inter-category video--image blends. The taxonomy is extensible to custom categories.}
  \label{fig:dataset_construction}
  \vspace{-10pt}
\end{figure*}

\subsection{Evaluation}
\label{sec:evaluation}
We evaluate \textit{video semantic mixing} along three distinct axes: directional concept alignment, temporal coherence, and video quality.
We report LPIPS-T, FVD, and ImageReward but exclude full-frame structural metrics such as SSIM, which penalize intended object-level changes.
Because absolute-difference blend scores can reward degenerate outputs (Sec.~\ref{app:clipbs}), we introduce \metricname\ (\metricnameabbr), a CLIP-based metric that measures the fused video's alignment shift relative to the source prompt and reference image.
Let $V_{\text{src}}$, $V_{\text{fused}}$, and $I_{\text{ref}}$ denote the source video, fused video, and reference image, respectively; let $E(\cdot)$ be the CLIP visual encoder; and let $L_{\text{src}}$ be the source-prompt embedding.
We compute:

\begin{equation}
\begin{aligned}
\text{CLIP-L}_{\text{src}} &= \text{sim}(E(V_{\text{src}}), L_{\text{src}}), \quad
\text{CLIP-I}_{\text{src}} = \text{sim}(E(V_{\text{src}}), E(I_{\text{ref}})), \\
\text{CLIP-L}_{\text{fused}} &= \text{sim}(E(V_{\text{fused}}), L_{\text{src}}), \quad
\text{CLIP-I}_{\text{fused}} = \text{sim}(E(V_{\text{fused}}), E(I_{\text{ref}})).
\end{aligned}
\end{equation}

We define the unnormalized image- and language-alignment changes as $\Delta_I$ and $\Delta_L$, respectively:
\begin{equation}
\Delta_I = \text{CLIP-I}_{\text{fused}}-\text{CLIP-I}_{\text{src}},\quad
\Delta_L = \text{CLIP-L}_{\text{fused}}-\text{CLIP-L}_{\text{src}},\quad
\text{CASS}=\Delta_I-\Delta_L.
\end{equation}
CTVB includes semantically close intra-category pairs, such as cow and sheep, and more distant inter-category pairs, such as astronaut and cat.
Consequently, the pre-edit CLIP alignment between the source video and reference image, $\text{CLIP-I}_{\text{src}}$, can vary substantially across examples.
Because \metricnameabbr\ uses the unnormalized changes $\Delta_I$ and $\Delta_L$, an equal change can represent a different relative shift depending on the corresponding pre-edit alignment.
We therefore normalize each change by its pre-edit CLIP score:
\begin{equation}
\begin{aligned}
\text{relCLIP-I} &= \frac{\Delta_I}{\text{CLIP-I}_{\text{src}}}, \qquad
\text{relCLIP-L} = \frac{\Delta_L}{\text{CLIP-L}_{\text{src}}}, \qquad
\text{rel-CASS} &= \text{relCLIP-I}-\text{relCLIP-L}.
\end{aligned}
\end{equation}
This normalization reduces sensitivity to differences in pre-edit alignment across examples.
For both metrics, higher values indicate a larger shift toward the reference and away from the source prompt.
Neither metric alone establishes balanced hybridization, source-identity retention, background preservation, or video quality; we assess those properties separately with qualitative, human, temporal, and quality evaluations.

\subsection{Baseline Comparisons}
\label{sec:baseline}

We evaluate the training-free methods FreeBlend~\citep{zhou2025freeblend}, adapted using per-frame editing and DynamiCrafter~\citep{xing2023dynamicrafter}; RAVE~\citep{kara2023raverandomizednoiseshuffling}, based on latent shuffling; and AnyV2V~\citep{ku2024anyv2vtuningfreeframeworkvideotovideo}, which propagates first-frame edits.
We also include AnimateDiffV2V~\citep{guo2023animatediff} for motion-preserving prompt-based editing and TokenFlow~\citep{tokenflow2023} with PnP and SDEdit configurations.

\begin{table}[h]
\centering
\small
\caption{\textbf{Quantitative comparison.} We report directional concept alignment (CASS, rel-CASS), temporal coherence (LPIPS-T), and video quality (FVD, ImageReward). Best in \textbf{bold}, second-best \underline{underlined}. Full-frame structural metrics are excluded, as the task deliberately alters the target object.}
\vspace{-0.2cm}
\label{tab:eval_comparison}
\resizebox{\linewidth}{!}{
\begin{tabular}{ll cc c cc}
\toprule
& \textbf{Method}
& \textbf{CASS}~$\uparrow$ & \textbf{rel-CASS}~$\uparrow$
& \textbf{LPIPS-T}~$\downarrow$
& \textbf{FVD}~$\downarrow$ & \textbf{ImageReward}~$\uparrow$\\
\cmidrule(lr){3-4}\cmidrule(lr){5-5}\cmidrule(lr){6-7}
& & \multicolumn{2}{c}{\footnotesize Directional alignment}
  & {\footnotesize Temporal}
  & \multicolumn{2}{c}{\footnotesize Video quality}\\
\midrule
\multicolumn{7}{l}{\textbf{\small Pretrained}} \\
& AnimateDiffV2V            & 0.68 & -0.41 & \textbf{0.009} & \textbf{1266} & \underline{0.246} \\
& TokenFlow PnP             & 2.87 & 0.02 & \underline{0.010} & 4580 & -2.087 \\
& TokenFlow SDEdit          & 1.98 & 0.05 & 0.150 & 4417 & -1.431 \\
\midrule
\multicolumn{7}{l}{\textbf{\small Training-free}} \\
& FreeBlend + DynamiCrafter & 1.47 & 0.01 & 0.016 & 6313 & -2.060 \\
& RAVE                      & \underline{3.80} & \underline{0.11} & 0.040 & 3546 & -0.721 \\
& AnyV2V                    & 2.31 & 0.08 & 0.020 & 3971 & -1.240 \\
\midrule
& \textbf{MoCA-Video (ours)} & \textbf{7.15} & \textbf{0.14} & 0.111 & \underline{3520} & \textbf{0.263} \\
\bottomrule
\end{tabular}}
\vspace{-10pt}
\end{table}

\begin{figure*}[h]
  \centering
  \includegraphics[width=0.95\linewidth]{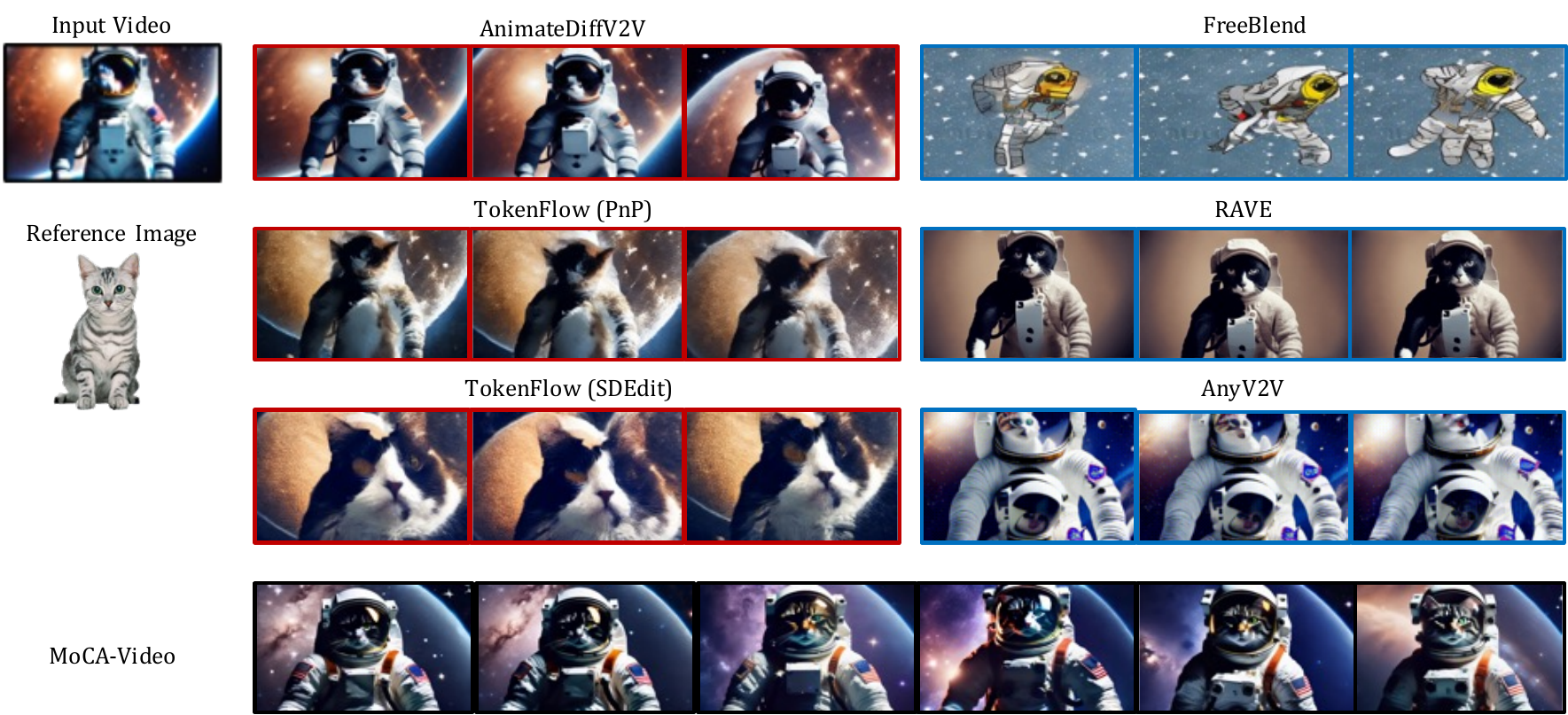}
  \caption{\textbf{Qualitative comparison of video semantic mixing.} In this astronaut--cat example, \modelnameabbr\ forms a visible hybrid: feline features persist across the displayed frames while source spacesuit and scene elements remain. AnimateDiffV2V largely preserves the source, whereas several stronger edits overwrite its structure or introduce artifacts. Red and blue borders mark pretrained and training-free baselines; full sequences appear in the Supplementary Material.}

  \label{fig:baseline_comparison}
\end{figure*}

Tab.~\ref{tab:eval_comparison} highlights a trade-off between directional concept shift and conventional temporal and quality metrics.
AnimateDiffV2V achieves the lowest FVD (1266) and LPIPS-T (0.009) but has the lowest CASS (0.68).
In the qualitative example in Fig.~\ref{fig:baseline_comparison}, it largely reproduces the original astronaut.

\noindent\textbf{Training-free methods over-edit.} FreeBlend, RAVE, and AnyV2V display a stronger reference signal than AnimateDiffV2V in Fig.~\ref{fig:baseline_comparison}, together with visible artifacts or source changes.
FreeBlend records the highest FVD (6313) and an ImageReward of $-2.06$, while RAVE attains the highest CASS among baselines (3.80).
The editing baselines have negative ImageReward scores ($-0.72$ to $-2.09$), whereas \modelnameabbr\ and AnimateDiffV2V have positive scores.

\noindent\textbf{\modelnameabbr\ emphasizes reference integration.} In Fig.~\ref{fig:baseline_comparison}, \modelnameabbr\ produces an astronaut--cat hybrid that retains visible elements of the spacesuit and scene while carrying feline appearance across the displayed frames.
Quantitatively, it achieves the highest CASS (7.15) and ImageReward (0.263) among the evaluated methods.
Its LPIPS-T (0.111) and FVD (3520) remain worse than the best values on those metrics, so the results should be read as a trade-off rather than uniform superiority.

\subsection{Ablation Studies}
\label{sec:ablation}
We ablate \modelnameabbr\ in two parts.
Component ablations disable each mechanism (Tab.~\ref{tab:abl_components}), and hyperparameter sweeps vary each parameter around the default $\lambda=0.8$, $\tau=0.4$, $c=2.0$ (Tab.~\ref{tab:abl_hparams}).
\begin{table}[t]
\centering
\small
\caption{\textbf{Component ablations.} Each row disables one mechanism of \modelnameabbr, on the full evaluation set. Defaults: $\lambda=0.8$, $\tau=0.4$, $c=2.0$. Best in \textbf{bold}, second-best \underline{underlined}.}
\label{tab:abl_components}
\resizebox{\linewidth}{!}{
\begin{tabular}{l cc c cc}
\toprule
Variant & CASS~$\uparrow$ & rel-CASS~$\uparrow$ & LPIPS-T~$\downarrow$ & FVD~$\downarrow$ & ImageReward~$\uparrow$ \\
\midrule
MoCA-Video (ours)           & \textbf{7.15} & \textbf{0.135} & \underline{0.111} & \underline{3520} & \underline{0.263} \\
\quad w/o injection         & 4.74 & 0.090 & 0.127 & \textbf{3447} & 0.256 \\
\quad w/o momentum          & \underline{5.98} & \underline{0.108} & 0.119 & 3602 & \textbf{0.324} \\
\quad w/o concept attention & 5.25 & 0.098 & \textbf{0.108} & 3670 & \textbf{0.324} \\
\bottomrule
\end{tabular}}
\end{table}

Tab.~\ref{tab:abl_components} shows that all three mechanisms increase CASS in this ablation.
Removing latent injection causes the largest drop, from 7.15 to 4.74.
The results use the default hyperparameters detailed in the Supplementary Material.

Removing \textit{concept-attention localization} yields CASS 5.25.
This variant replaces the localized mask with full-image injection, so the reference is no longer concentrated on the target region.

Disabling \textit{momentum propagation} reduces CASS from 7.15 to 5.98, which is consistent with its intended role in carrying the injected concept across frames.
LPIPS-T changes only from 0.119 without momentum to 0.111 with momentum, so the ablation supports a modest temporal-coherence improvement rather than a large stabilization effect.
Disabling momentum or localization raises ImageReward from 0.263 to 0.324, revealing a trade-off between the measured directional concept shift and per-frame reward.
The full method's ImageReward nevertheless remains above all evaluated baselines in Tab.~\ref{tab:eval_comparison}.

\noindent\textbf{Hyperparameter sensitivity.} We vary each hyperparameter on a fixed subset (5 prompts $\times$ 2 concepts), holding the others at their defaults.
Absolute CASS values are not comparable to Tab.~\ref{tab:eval_comparison}; Tab.~\ref{tab:abl_hparams} supports comparisons only within each sweep on this subset.
CASS is highest at $\lambda=0.8$ and $\tau=0.4$ among the tested values.
The tested momentum clips yield the highest CASS at $c=2.0$.
We therefore use $(\lambda,\tau,c)=(0.8,0.4,2.0)$ as the experimental defaults.

\begin{table}[h]
\centering
\small
\caption{Hyperparameter ablation (CASS).
Each parameter is varied while the others remain fixed at their
default values ($^\ast$). Among the tested values, CASS is highest at $\lambda=0.8$, $\tau=0.4$, and $c=2.0$.}
\label{tab:abl_hparams}
\begin{tabular}{@{}lcccccc@{}}
\toprule
$\lambda$ & 0.5 & \textbf{0.8$^\ast$} & 1.0 & 1.5 & 2.0 \\
CASS & 2.69 & \textbf{4.09} & 1.02 & 0.54 & 0.40 \\
\midrule
$\tau$ & 0.1 & 0.2 & 0.3 & \textbf{0.4$^\ast$} & 0.5 \\
CASS & 0.89 & 1.60 & 1.56 & \textbf{4.09} & 2.37 \\
\midrule
$c$ & off & 0.5 & 1.0 & \textbf{2.0$^\ast$} & 5.0 \\
CASS & 0.64 & 0.69 & 1.33 & \textbf{4.09} & 0.64 \\
\bottomrule
\end{tabular}
\end{table}

\section{Conclusion}
\label{sec:conclusion}
We presented \modelnameabbr, a training-free framework for video semantic mixing.
The method operates in the latent trajectory of a frozen video diffusion model: it localizes the target with the model's cross-attention, injects the reference concept into that region, and carries the injected prediction across frames through momentum-based correction.
We also introduced CTVB, a benchmark with 21 base videos and 106 intra- and inter-category video--image pairs that combines \metricnameabbr/rel-\metricnameabbr, LPIPS-T, FVD, ImageReward, qualitative comparisons, and human evaluation.
On the evaluated baselines, \modelnameabbr\ achieves the highest CASS, rel-CASS, and ImageReward, while its LPIPS-T and FVD results expose a trade-off between reference integration and conventional temporal and video-quality metrics.
Our UNet-only experiments and the limits of CASS bound this evidence, but the results support concept-localized intervention in continuous video-diffusion trajectories as a direction for inference-time controllable generation.

{\small
\bibliographystyle{plainnat}
\bibliography{main}
}

\clearpage
\appendix

\section{Supplementary Material}
\label{sec:app}

\subsection{Algorithm}

Algorithm~\ref{alg:momentum_unified} details momentum-based propagation along the diagonal schedule. We use $t$ for diffusion timestep and $f$ for frame index.
Each iteration applies momentum correction and the DDIM update, injects the reference latent during low-noise steps, and retains the resulting prediction to guide the next frame.

\begin{algorithm}[h]
    \caption{Momentum-based concept propagation}
    \label{alg:momentum_unified}
    \begin{algorithmic}[1]
        \REQUIRE Noisy frame latents $x_{t,f}$,
        binary masks $a_{t,f}$,
        reference latent $z^{\mathrm{ref}}$,
        conditioning strength $\lambda$,
        momentum decay $\beta$,
        direction scale $\eta$,
        clip value $c$,
        and schedule parameters $\kappa_0,T$

        \STATE $v_{t,0} \gets 0$
        \STATE $\tilde{x}_{0,0}^{(t)} \gets \varnothing$

        \FOR{each frame $f$ at diffusion timestep $t$ along the diagonal}

            \STATE
            $\hat{x}_{0,f}^{(t)}
            \gets
            \dfrac{
                x_{t,f}
                -
                \sqrt{1-\alpha_t}\,
                \epsilon_\theta(x_{t,f},t)
            }{
                \sqrt{\alpha_t}
            }$
            \COMMENT{raw clean prediction}

            \STATE
            $d_{t,f}
            \gets
            \sqrt{1-\alpha_{t-1}-\sigma_t^2}\,
            \epsilon_\theta(x_{t,f},t)$

            \IF{$\tilde{x}_{0,f-1}^{(t)} \neq \varnothing$}

                \STATE
                $g_{t,f}
                \gets
                \bigl(
                    \hat{x}_{0,f}^{(t)}
                    -
                    \tilde{x}_{0,f-1}^{(t)}
                \bigr)
                +
                \eta\,d_{t,f}$

                \STATE
                $v_{t,f}
                \gets
                \mathrm{clip}\!\left(
                    \beta v_{t,f-1}
                    +
                    (1-\beta)g_{t,f},
                    -c,
                    c
                \right)$

                \STATE
                $\kappa_t
                \gets
                \kappa_0(1-t/T)$

                \STATE
                $\hat{x}_{0,f}^{\mathrm{(corr)},(t)}
                \gets
                \hat{x}_{0,f}^{(t)}
                +
                \kappa_t v_{t,f}$
                \COMMENT{momentum correction}

            \ELSE

                \STATE
                $v_{t,f} \gets 0$

                \STATE
                $\hat{x}_{0,f}^{\mathrm{(corr)},(t)}
                \gets
                \hat{x}_{0,f}^{(t)}$

            \ENDIF

            \STATE
            $x_{t-1,f}
            \gets
            \sqrt{\alpha_{t-1}}\,
            \hat{x}_{0,f}^{\mathrm{(corr)},(t)}
            +
            d_{t,f}
            +
            \sigma_t\epsilon_{t,f}$
            \COMMENT{output latent}

            \IF{$t \leq t'$}

                \STATE
                $\tilde{x}_{0,f}^{(t)}
                \gets
                (1-a_{t,f})
                \odot
                \hat{x}_{0,f}^{\mathrm{(corr)},(t)}
                +
                a_{t,f}
                \odot
                \bigl(\lambda z^{\mathrm{ref}}\bigr)$
                \COMMENT{inject reference}

            \ELSE

                \STATE
                $\tilde{x}_{0,f}^{(t)}
                \gets
                \hat{x}_{0,f}^{\mathrm{(corr)},(t)}$

            \ENDIF

            \STATE
            Retain $\tilde{x}_{0,f}^{(t)}$ and $v_{t,f}$
            for the next frame

        \ENDFOR

        \STATE \textbf{return} output latents
    \end{algorithmic}
\end{algorithm}

\noindent\textbf{Analysis.}
For a fixed diffusion timestep $t$, the buffer evolves across
frames according to
\[
v_{t,f}
=
\mathrm{clip}\!\left(
\beta v_{t,f-1}+(1-\beta)g_{t,f},
-c,c
\right).
\]
When clipping is inactive, this recurrence is a first-order
low-pass filter over the frame index $f$, with an effective
averaging window of approximately $1/(1-\beta)$ frames
($\approx 10$ for $\beta=0.9$).
The modest LPIPS-T decrease from $0.119$ without momentum to
$0.111$ with momentum is consistent with this filtering
interpretation.
Clipping guarantees $\lVert v_{t,f}\rVert_\infty\leq c$, so each
correction satisfies
$\lVert\kappa_t v_{t,f}\rVert_\infty\leq\kappa_0c$.
This per-step bound does not by itself establish output quality
or temporal stability.
The schedule $\kappa_t=\kappa_0(1-t/T)$ reduces the correction at
high noise and increases it as denoising approaches $t=0$.

\subsection{Additional Visual Results}
\label{app:visuals}
\begin{figure*}[h]
  \centering
  \includegraphics[width=\linewidth]{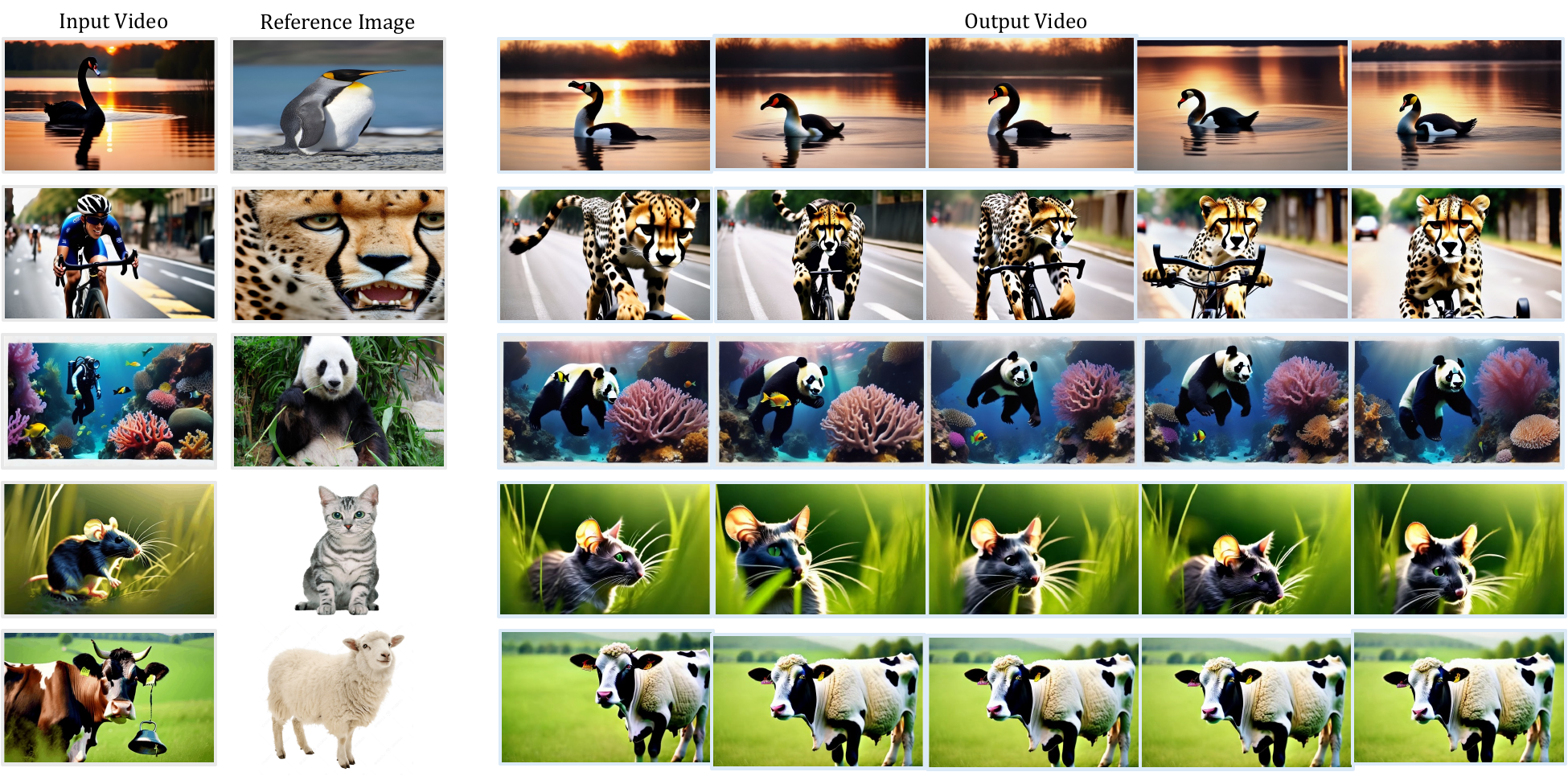}
  \caption{\textbf{Additional qualitative results.} Examples across diverse concept pairs illustrate how source and reference appearances blend.}
  \vspace{-15pt}
  \label{fig:supp_qualitative}
\end{figure*}

We provide additional qualitative comparisons on concept-blending tasks in Fig.~\ref{fig:supp_qualitative}.
We use the UNet-based VideoCrafter2 backbone for a backbone-consistent comparison with FreeBlend, TokenFlow, RAVE, AnimateDiffV2V, and AnyV2V.
\subsection{Additional Qualitative Results on DAVIS Pairs}
\label{app:davis_results}
\begin{figure*}[h]
  \centering
  \includegraphics[width=\textwidth]{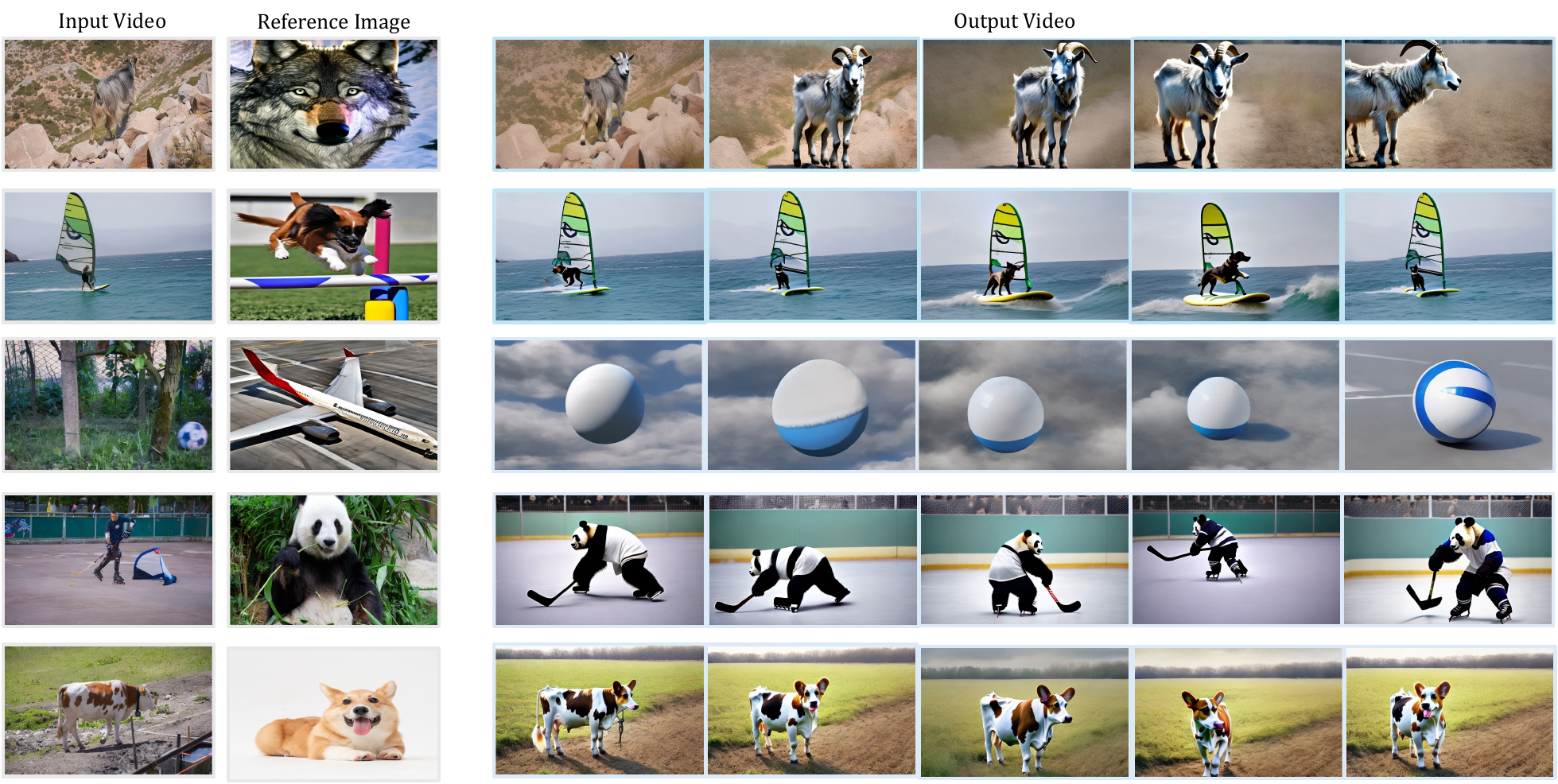}
  \caption{\textbf{Additional qualitative results on DAVIS-derived pairs.} Each row shows an input-video frame (left), reference image (center), and frames from the \modelnameabbr\ output (right). The examples span intra- and inter-category blends across animals, objects, and human activities.}
  \label{fig:davis_qualitative}
\end{figure*}

Fig.~\ref{fig:davis_qualitative} shows additional \modelnameabbr\ results on DAVIS-16 entity pairs spanning intra- and inter-category blends.
The displayed frames include goat--wolf, windsurfer--dog, soccer-ball--airplane, roller-hockey-player--panda, and cow--corgi pairs.
They qualitatively illustrate reference integration together with source layout and scene content; temporal claims require the corresponding videos.

\paragraph{Mask source.} \modelnameabbr\ localizes the target object from the model's cross-attention by default and requires no external supervision.
The localization stage is modular: it consumes a binary mask $a_t$ and is agnostic to how that mask is produced.
For DAVIS-derived videos, which provide ground-truth segmentation masks, the provided annotation can replace the attention-derived mask while the injection and momentum stages remain unchanged.
This replacement is optional; the default method uses attention-based localization.

\subsection{Why CASS? Failure Modes of Absolute-Difference Blend Scores}
\label{app:clipbs}
A semantic-mixing metric should distinguish a directional shift from the source toward the reference from a large but uninformative score change.
FreeBlend's CLIP-BS~\citep{zhou2025freeblend} does not encode this direction because it takes an absolute difference between aggregate similarities.
We analyze two resulting failure modes and use them to motivate CASS.
FreeBlend evaluates mixing with a blend score, CLIP-BS, defined as the absolute difference between the fused video's combined similarity to the two source prompts and the corresponding similarity of the originals:
\begin{equation}
\begin{aligned}
s_{\mathrm{mix}}  &= \mathrm{sim}(V_{\mathrm{fused}}, t_A) + \mathrm{sim}(V_{\mathrm{fused}}, t_B),\\
s_{\mathrm{orig}} &= \mathrm{sim}(V_{A}, t_A) + \mathrm{sim}(V_{B}, t_B),\\
\mathrm{CLIP\text{-}BS} &= \bigl| s_{\mathrm{mix}} - s_{\mathrm{orig}} \bigr|.
\end{aligned}
\end{equation}
Because CLIP-BS scores similarity to each prompt independently and then takes an absolute difference, it can assign high values to outputs that fail the blending objective.
We identify two failure modes.
\emph{(i) One-sided dominance.} A fused video that strongly resembles only one concept (say $A$) raises $\mathrm{sim}(V_{\mathrm{fused}}, t_A)$ while ignoring $B$, yet can still yield a large $s_{\mathrm{mix}}$ and CLIP-BS.
The same issue arises if both concepts appear as separate, side-by-side objects rather than a hybrid.
\emph{(ii) Collapse.} If $s_{\mathrm{mix}}$ becomes very small because the output resembles neither concept, the absolute value can turn the large negative difference $s_{\mathrm{mix}} - s_{\mathrm{orig}}$ into a high score.
CLIP-BS therefore does not by itself establish that the output blends the two concepts.

\begin{table}[h]
\centering
\small
\caption{FreeBlend's blend metrics on 100 test samples. FreeBlend scores higher on both metrics, whereas Fig.~\ref{fig:baseline_comparison} and Tab.~\ref{tab:eval_comparison} provide separate qualitative and video-quality comparisons.}
\label{tab:clipbs}
\begin{tabular}{@{}lcc@{}}
\toprule
Method & CLIP-BS$\uparrow$ & DINO-BS$\uparrow$ \\
\midrule
FreeBlend   & 6.65 & 0.27 \\
MoCA-Video  & 4.00 & 0.12 \\
\bottomrule
\end{tabular}
\end{table}

On 100 test samples, FreeBlend obtains higher CLIP-BS and DINO-BS than \modelnameabbr\ (Tab.~\ref{tab:clipbs}).
DINO-BS is the analogous blend score computed with DINO features instead of CLIP.
In the shown qualitative comparison, FreeBlend distorts the source (Fig.~\ref{fig:baseline_comparison}) and records the highest FVD among the evaluated methods (Tab.~\ref{tab:eval_comparison}).
This mismatch illustrates that CLIP-BS and DINO-BS need not track the desired output properties.
CASS avoids the absolute-difference collapse mode by measuring a \emph{directional} shift toward the reference and away from the source prompt.
However, CASS alone does not distinguish balanced hybridization from reference-dominant replacement, so we interpret it alongside qualitative, human, temporal, and video-quality evaluations.

\subsection{Human Evaluation Protocol}
\label{app:human_eval}
\subsubsection{User Study}
We recruited twenty volunteers (aged 18--45, balanced gender) from our university community, all of whom reported normal or corrected-to-normal vision and no prior involvement in this project.
Each participant completed eight independent trials in a single session lasting approximately twenty minutes.
Participants provided electronic consent and reviewed a demonstration trial before the study.
In each trial, participants first viewed an input video and a reference image showing two concepts.
They then viewed three anonymized two-second clips generated by \modelnameabbr, AnimateDiffV2V, and FreeBlend+DynamiCrafter in random order.
Participants rated each clip on four dimensions using a 1--5 Likert scale: Blending Quality, Video Consistency, Character Consistency, and Overall Quality.
The interface used a multiple-choice grid so that participants scored every clip on each criterion.
\begin{figure}[h]
    \centering
    \includegraphics[width=1.0\linewidth]{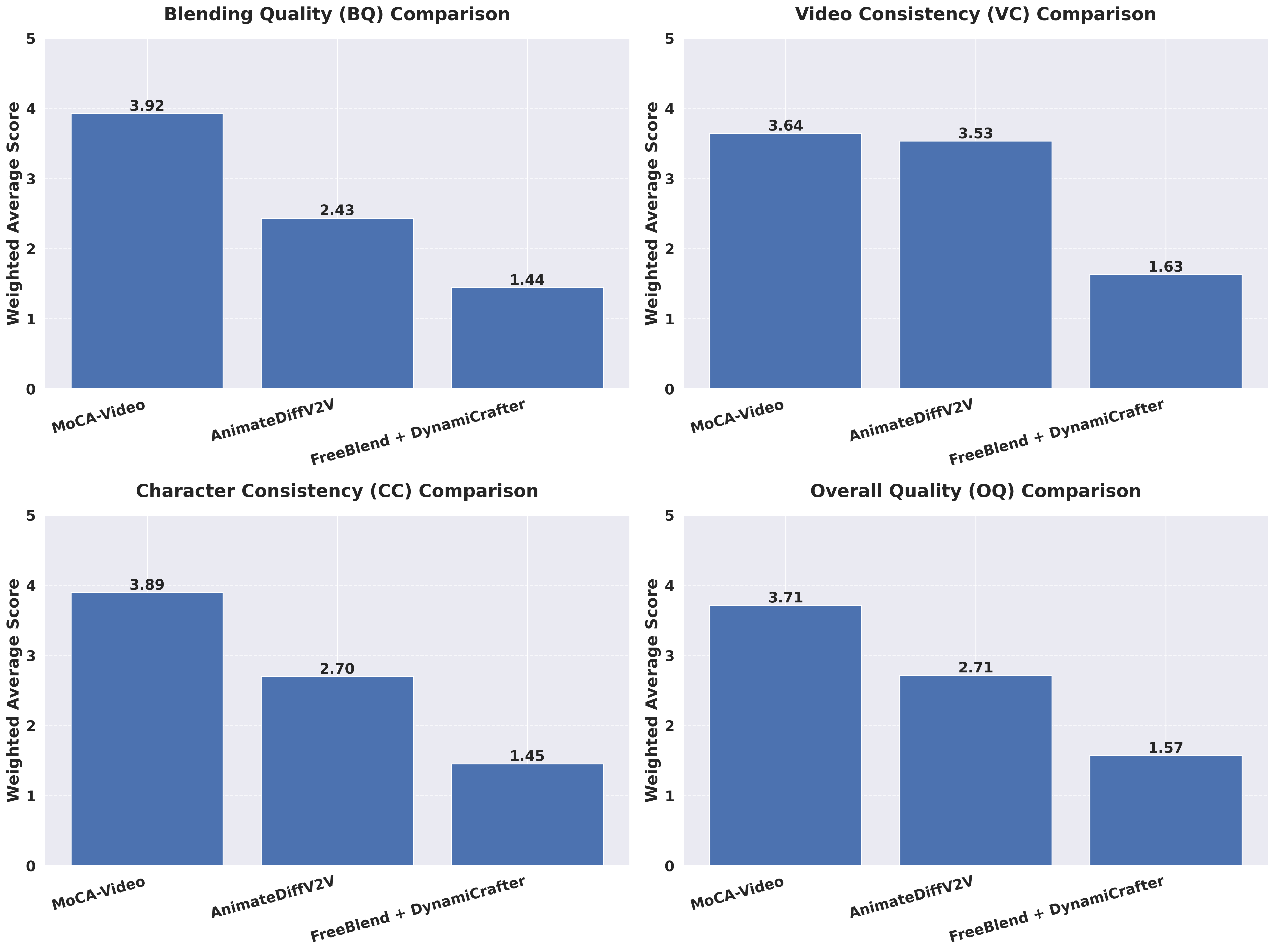}
    \caption{Mean ratings from 20 participants, each completing eight trials, for MoCA-Video, AnimateDiffV2V, and FreeBlend+DynamiCrafter. MoCA-Video receives the highest mean Blending Quality and Overall Quality ratings; Video Consistency and Character Consistency are reported as separate subjective measures.
}
    \label{fig:user study comparison}
\end{figure}

Figure~\ref{fig:user study comparison} reports subjective judgments of blending quality, motion coherence, character consistency, and overall quality.
The study complements the automated metrics but does not test correlation with CASS or rel-CASS.

\subsection{Experimental Setup}
\label{app:exp_setup}
We run all experiments with the UNet-based VideoCrafter2 backbone, using Python 3.10, PyTorch 2.1, and CUDA 11.8 on a single NVIDIA A100 GPU (FP16).
We sample frames from the second half of each generated sequence, around $\lfloor \texttt{video\_length}/2 \rfloor$, to evaluate outputs after semantic injection has accumulated.
We use injection timestep $t' = 300$, conditioning strength $\lambda = 0.8$, localization threshold $\tau = 0.4$, direction weight $\eta = 0.05$, momentum decay $\beta = 0.9$, and momentum clip $c = 2.0$ with schedule $\kappa_0 = 0.1$.
Remaining scheduler and denoising-step settings use their defaults.

\subsubsection{Computational Efficiency}
Table~\ref{tab:inference_time} compares per-frame inference time on the same A100 hardware, excluding preprocessing such as model initialization and base-video generation.
\modelnameabbr\ requires about 3.2\,s per output frame, similar to AnimateDiffV2V (3.8\,s), RAVE (3.0\,s), and the DynamiCrafter-only portion of FreeBlend + DynamiCrafter (3.3\,s).
It is slower than TokenFlow SDEdit (1.9\,s) and faster than TokenFlow PnP (9.4\,s).
Because the table does not isolate attention-localization time from the diagonal scheduler, it does not quantify localization overhead separately.
The methods also produce different sequence lengths, so the per-frame measurements should not be read as total-runtime comparisons.

\begin{table}[h]
\centering
\small
\caption{Per-frame inference time on a single A100 (FP16), excluding preprocessing. Sequence lengths differ, and FreeBlend's image-mixing time is excluded from its row.}
\label{tab:inference_time}
\begin{tabular}{@{}lcc@{}}
\toprule
\textbf{Method} & \textbf{Frames} & \textbf{Time (s/frame)} \\
\midrule
AnimateDiffV2V                          & 32 & 3.8 \\
FreeBlend + DynamiCrafter\textsuperscript{a} & 16 & 3.3 \\
TokenFlow (PnP)                         & 16 & 9.4 \\
TokenFlow (SDEdit)                      & 16 & 1.9 \\
RAVE                                    & 16 & 3.0 \\
\midrule
\textbf{MoCA-Video (ours)}              & 50 & \textbf{3.2} \\
\bottomrule
\end{tabular}

\vspace{2pt}
\raggedright\footnotesize
\textsuperscript{a}3.3\,s is the animation time (DynamiCrafter); FreeBlend's image-level mixing is additional.
\end{table}

\subsection{Diagonal Denoising Scheduler Implementation}

We summarize the diagonal denoising scheduler introduced by FIFO-Diffusion~\citep{kim2024fifo} and explain how \modelnameabbr\ inserts semantic mixing into it.

\textbf{Queue Structure.} The diagonal denoising scheduler processes video frames in a queue:
\begin{equation}
    Q = \{z_1^{\tau_1}, z_2^{\tau_2}, \ldots, z_{nf}^{\tau_{nf}}\},
\end{equation}
where each frame $i$ is at a different noise level $\tau_i$ with $0 < \tau_1 < \tau_2 < \ldots < \tau_{nf} = T$.
Unlike standard parallel denoising where all frames share the same timestep, this diagonal arrangement enables frames to reference cleaner (lower noise) neighbors during denoising.

\textbf{Partitioned Denoising.} At each iteration, the queue is partitioned into $n$ blocks of size $f$ (the base model's temporal capacity window):
\begin{equation}
    Q = [Q_0, Q_1, \ldots, Q_{n-1}],
\end{equation}
where each block $Q_k$ is denoised via:
\begin{equation}
    Q_k \leftarrow \Phi(Q_k, \tau_k, c; \epsilon_\theta),
\end{equation}
with $\tau_k = \{\tau_{kf+1}, \ldots, \tau_{(k+1)f}\}$, $c$ represents text condition, and $\Phi(\cdot)$ denoting the DDIM sampler.

\textbf{Semantic Injection.} Within the editing window $t \le t'$, we inject reference features on the predicted clean sample $\hat{x}_0$ inside the attention-derived region:
\begin{equation}
    \hat{x}_0' = (1 - a_t) \odot \hat{x}_0 + a_t \odot \bigl(\lambda\, z^{ref}\bigr),
\end{equation}
where $a_t$ is the binary localization mask obtained by thresholding the concept-attention map (\ref{eq:mask}), $z^{ref} = E(I^{ref})$ is the encoded reference, and $\lambda$ is the conditioning strength.
The mask is recomputed from the denoiser's own cross-attention at each step, requiring no external segmentation or tracking.

\textbf{FIFO Queue Management.} After each iteration:
\begin{enumerate}
    \item \textbf{Dequeue}: The fully denoised frame $z_1^{\tau_0}$ at the queue head is removed and output
    \item \textbf{Enqueue}: A new noisy latent $z_{i+nf}^{\tau_{nf}}$, initialized using the dequeued frame, is added to the tail
    \item \textbf{Propagation}: All remaining frames shift forward, creating a sliding temporal window
\end{enumerate}

\textbf{Temporal Coherence Mechanism.} This diagonal pattern enables temporal propagation of semantic edits.
As frame $i$ moves through the queue from high noise ($\tau_{nf}$) to clean ($\tau_0$), it observes already-blended neighboring frames.
Each frame references both \emph{cleaner} preceding frames and \emph{noisier} following frames during denoising.

\textbf{Difference from Standard Denoising.} Traditional window-based denoisers process all latents in a temporal window at the same timestep $t$, whereas the diagonal scheduler processes frames at different noise levels $\{\tau_1, \ldots, \tau_{nf}\}$.
This heterogeneous structure lets cleaner frames provide reference points for noisier frames undergoing semantic transformation.

\subsection{Prompts for Qualitative Results}
\label{app:prompts}
Table~\ref{tab:qual_prompts} lists the source and conditioning prompts used for the qualitative \modelnameabbr\ results on our dataset (Figs.~\ref{fig:semantic_mixing_examples}, \ref{fig:baseline_comparison}, and~\ref{fig:supp_qualitative}).
Each entry pairs a source prompt used to generate the base video with a conditioning concept blended into the target entity.

\begin{table}[h]
\centering
\small
\caption{Source and conditioning prompts for the qualitative results.}
\label{tab:qual_prompts}
\setlength{\tabcolsep}{5pt}
\begin{tabularx}{\linewidth}{@{}Xl@{}}
\toprule
\textbf{Source prompt} & \textbf{Condition} \\
\midrule
A surfer riding a wave at sunset & panda \\
A superhero in a dynamic pose against a city skyline, cape flowing in the wind & deer \\
A majestic eagle soaring through the sky with its wings spread wide & corgi \\
A horse galloping across an open meadow & unicorn \\
A firefighter in full gear, standing in front of a burning building & bear \\
An astronaut floating in space against a starry background, wearing a detailed white spacesuit & cat \\
A black swan gliding gracefully across a serene lake at sunset & penguin \\
A cyclist racing down a city street & cheetah \\
A deep-sea diver exploring a coral reef, surrounded by colorful fish & panda \\
A grey mouse eating in a grassland & cat \\
A cow with a bell grazing in a green pasture & sheep \\
\bottomrule
\end{tabularx}
\end{table}

Table~\ref{tab:davis_prompts} lists the prompts used for the DAVIS-derived qualitative results in Fig.~\ref{fig:davis_qualitative}.
These examples use short descriptions based on the DAVIS-16 class labels to reduce elaborate textual conditioning.
The examples illustrate results under minimal source prompts; isolating the relative influence of the prompt and reference would require a controlled prompt ablation.

\begin{table}[h]
\centering
\small
\caption{Source and conditioning prompts for the DAVIS qualitative results. Source prompts are intentionally minimal to isolate the effect of visual guidance.}
\label{tab:davis_prompts}
\setlength{\tabcolsep}{5pt}
\begin{tabularx}{\linewidth}{@{}Xl@{}}
\toprule
\textbf{Source prompt} & \textbf{Condition} \\
\midrule
A goat walking on rocks & wolf \\
A man windsurfing with the sail & dog \\
A rolling soccer ball & airplane \\
A man wearing hockey skates & panda \\
A cow with a bell around its neck & corgi \\
\bottomrule
\end{tabularx}
\end{table}

\subsection{Code and Videos}
\label{app:code}
Project repository: \url{https://anonymous.4open.science/r/MoCA-Video-0058}.

\subsection{Use of Large Language Models}
We used commercial LLMs (ChatGPT and Claude) for editorial polishing and for generating example prompts in our evaluation dataset.
Their assistance was limited to improving writing clarity and academic style and to producing diverse sample prompts.
All technical contributions, experimental results, and scientific conclusions are the authors' own, and all AI-assisted content was reviewed and validated by the authors.

\end{document}